\def\year{2021}\relax
\documentclass[letterpaper]{article} %
\usepackage{aaai21}  %
\usepackage{times}  %
\usepackage{helvet} %
\usepackage{courier}  %
\usepackage[hyphens]{url}  %
\usepackage{graphicx} %
\usepackage{natbib}  %
\usepackage{caption} %
\usepackage{amssymb}
\usepackage{enumitem}
\usepackage{listings}
\usepackage{xstring}
\usepackage{graphicx}
\usepackage{pbox}
\usepackage{subcaption}
\usepackage{epstopdf}
\usepackage{xstring}

\usepackage{balance}
\usepackage{pbox}
\usepackage{multirow}
\usepackage{enumitem}

\usepackage{multicol}

\usepackage{tabularx}
\usepackage{tcolorbox}
\usepackage{color, colortbl}
\usepackage[show]{inotes}
\usepackage{soul}
\usepackage{booktabs}
\usepackage{multirow}
\usepackage[switch]{lineno}

\usepackage{comment}

\title{HumAID: Human-Annotated Disaster Incidents Data \\from Twitter with Deep Learning Benchmarks}

\author{Firoj Alam, Umair Qazi, Muhammad Imran, Ferda Ofli\\}
\affiliations{
    Qatar Computing Research Institute, HBKU, Qatar\\
    \{fialam,uqazi,mimran,fofli\}@hbku.edu.qa
}

\begin{document}
\maketitle

\begin{abstract}
Social networks are widely used for information consumption and dissemination, especially during time-critical events such as natural disasters. Despite its significantly large volume, social media content is often too noisy for direct use in any application. Therefore, it is important to filter, categorize, and concisely summarize the available content to facilitate effective consumption and decision-making. To address such issues automatic classification systems have been developed using supervised modeling approaches, thanks to the earlier efforts on creating labeled datasets. However, existing datasets are limited in different aspects (e.g., size, contains duplicates) and less suitable to support more advanced and data-hungry deep learning models. 
In this paper, we present a new large-scale dataset with ${\sim}$77K human-labeled tweets, sampled from a pool of ${\sim}$24 million tweets across 19 disaster events that happened between 2016 and 2019. Moreover, we propose a data collection and sampling pipeline, which is important for social media data sampling for human annotation. We report multiclass classification results using classic and deep learning (fastText and transformer) based models to set the ground for future studies. The dataset and associated resources are publicly available.\footnote{\url{https://crisisnlp.qcri.org/humaid_dataset.html}}
\end{abstract}

\section{Introduction}
\label{sec:introduction}
Recent studies highlight the importance of analyzing social media data during disaster events~\cite{imran2015processing} as it helps decision-makers to plan relief operations. However, most of the actionable information on social media is available in the early hours of a disaster when information from other traditional data sources is not available. However, utilizing this information requires time-critical analysis of social media streams for aiding humanitarian organizations, government agencies, and public administration authorities to make timely decisions and to launch relief efforts during emergency situations~\cite{starbird2010chatter,vieweg2010microblogging}. Among various social media platforms, Twitter has been widely used, on one hand, to disseminate information, and on the other, to collect, filter, and summarize information~\cite{alam2019descriptive}. 
As the volume of information on social media is extremely high~\cite{castillo2016big},
automated data processing is necessary to filter redundant and irrelevant content and categorize useful content. There are many challenges to dealing with such large data streams and extracting useful information. Those include parsing unstructured and brief content, filtering out irrelevant and noisy content, handling information overload, among others.

Typical approaches tackling this problem rely on supervised machine learning techniques, i.e., classify each incoming tweet into one or more of a pre-defined set of classes. In the past, several datasets for disaster-related tweets classification were published~\cite{olteanu2014crisislex,imran2016lrec,alam2018crisismmd}. These resources have supported NLP community to advance research and development in the \textit{crisis informatics}\footnote{\url{https://en.wikipedia.org/wiki/Disaster_informatics}} domain in many ways~\cite{purohit2013twitris,burel2018crisis,imran2014aidr,kumar2011tweettracker,okolloh2009ushahidi,Alam2019}.
Deep neural networks have shown SOTA performance in many NLP tasks and application areas. However, deep learning algorithms are usually data-hungry, whereas the existing datasets in the crisis informatics domain are limited in different respects, 
which restricts the development of more sophisticated deep learning models.

We have so far investigated the existing datasets to understand their limitations for future research. These limitations can be summarized as follows. The existing datasets cover small-scale events. They contain exact-or-near duplicate tweets (e.g., CrisisLexT26 \cite{olteanu2014crisislex}),\footnote{Existing datasets contains such duplicates for different reasons: retweeted tweet, and same tweet collected in different data collection.} which affects robustness of the trained models. They are usually dominated by tweets that come from outside of disaster hit areas and are usually about prayers and thoughts. We have also examined the existing literature to identify which categories are important for humanitarian organizations to extract actionable information and facilitate response efforts~\cite{nemeskey2018emergency,kropczynski2018identifying,strassel2017situational}. Such an analysis and understanding has motivated us to develop a new, large-scale dataset that can take crisis informatics research to the next level by affording the ability to develop more sophisticated models. 

Hence, in this paper, we present the largest publicly available human annotated Twitter dataset, called {\bf HumAID}: {\bf Hum}an-{\bf A}nnotated Disaster {\bf I}ncidents {\bf D}ata, for crisis informatics research. It has the following characteristics. {\em(i)} The dataset contains over 77,000 labeled tweets, which were sampled from 24 million tweets collected during 19 major real-world disasters that took place between 2016 and 2019, 
including hurricanes, earthquakes, wildfires, and floods. {\em (ii)} HumAID encompasses different disaster types across different time frames and locations. {\em(iii)} The dataset is more balanced in terms of disaster types and more consistent in terms of label agreement with regards to the existing datasets. {\em (iv)} Thanks to our carefully designed data filtering and sampling pipeline, HumAID consists of tweets that are more likely to be from the disaster-hit areas, and hence, contain more useful information coming from eyewitnesses or affected individuals. {\em(v)} Our annotation scheme consists of 11 categories representing critical information needs of a number of humanitarian organizations, including United Nations OCHA's needs reported in the MIRA framework\footnote{\url{https://www.humanitarianresponse.info/en/programme-cycle/space/document/mira-framework}} and previous studies \cite{vieweg2014integrating,olteanu2014crisislex,imran2016lrec,alam2018crisismmd,mccreadie2019trec}. {\em(vi)} Finally, HumAID is the largest dataset in comparison to the existing datasets in the crisis informatics domain.

Our focus was developing a large-scale human-labeled English tweets dataset covering several categories useful for humanitarian organizations during \textit{natural disasters}. 
To obtain such a large-scale dataset we used Amazon Mechanical Turk\footnote{\url{https://www.mturk.com/}} for the annotation. 
Furthermore, we used the labeled tweets to obtain benchmark results with both classical (i.e., SVM and RF) and deep learning algorithms (i.e., fastText and transformer based models). Our extensive experiments show that deep learning models outperform traditional supervised learning algorithms. Last but not least, we share dataset, code, and data splits with the research community for both reproducibility and further enhancements.

The rest of the paper is organized as follows. Section~\ref{sec:related_works} provides a brief overview of previous work. Section~\ref{sec:dataset} describes our data collection approaches, and Section~\ref{sec:manual_annotations} provides annotation procedures. We report experimental results in Section~\ref{sec:experimental_design} and discuss future research directions in Section~\ref{sec:discussion_future_works}. Finally, we conclude the paper in Section~\ref{sec:conclutions}.

\section{Related Work}
\label{sec:related_works}
Over the past few years, there has been a major research effort on analyzing social media content (mainly Twitter and Facebook) for humanitarian aid. Key challenges addressed in these studies include data filtering, classification, information extraction, and summarization to enhance situational awareness and mine actionable information~\cite{sakaki2010earthquake,imran2014aidr,saravanou2015twitter,tsou2017building,martinez2018twitter}. Most of these studies have been possible thanks to the publicly available datasets. 

\subsection{Existing Datasets}
\label{ssec:related-work-data}
Below we provide a brief overview of the existing datasets. 

\textit{\textbf{CrisisLex}} comprises two datasets, i.e., CrisisLexT26 and CrisisLexT6~\cite{olteanu2014crisislex}. 
The CrisisLexT26 dataset consists of $\sim$28,000 labeled tweets from 26 different disaster events that took place in 2012 and 2013. It includes disaster type and sub-type, and coarse- and fine-grained humanitarian class labels. 
CrisisLexT6 contains $\sim$60,000 labeled tweets from six disaster events that occurred between October 2012 and July 2013. 
Annotation of CrisisLexT6 includes \textit{related} vs.\ \textit{not-related}.

\textit{\textbf{CrisisMMD}} is a multimodal and multitask dataset consisting of $\sim$18,000 labeled tweets and associated images~\cite{alam2018crisismmd,ofli2020analysis}. Tweets have been collected from seven natural disaster events that took place in 2017. The annotations include three tasks: (i) \textit{informative} vs.\ \textit{not-informative}, (ii) humanitarian categories (eight classes), and (iii) damage severity levels (three classes). The third annotation task, i.e., damage severity (mild, severe and none), was applied only on images.

\textit{\textbf{CrisisNLP}} consists of ${\sim}$50,000 human-labeled tweets collected from 19 different disasters that happened between 2013 and 2015, and annotated according to different schemes including classes from humanitarian disaster response and some classes pertaining to health emergencies~\cite{imran2016lrec}.

\textit{\textbf{Disaster Response Data}} contains 30,000 tweets with 36 different categories, collected during disasters such as an earthquake in Haiti in 2010, an earthquake in Chile in 2010, floods in Pakistan in 2010, Hurricane Sandy in USA in 2012, and news articles.\footnote{\url{https://www.figure-eight.com/dataset/combined-disaster-response-data/}} %

\textit{\textbf{Disasters on Social Media}} dataset comprises 10,000 tweets annotated with labels \textit{related} vs.\ \textit{not-related} to the disasters.\footnote{\url{https://data.world/crowdflower/disasters-on-social-media}}

\textit{\textbf{SWDM2013}} consists of two data collections. The Joplin collection contains 4,400 labeled tweets collected during the tornado that struck Joplin, Missouri on May 22, 2011. The Sandy collection contains 2,000 labeled tweets collected during Hurricane Sandy, that hit Northeastern US on Oct 29, 2012~\cite{imran2013practical}.

\textit{\textbf{Eyewitness Messages}} dataset contains ${\sim}$14,000 tweets with labels (i) direct-eyewitness, (ii) indirect-eyewitness, (iii) non-eyewitness, and (iv) don't know, for different event types such as flood, earthquake, fire, and hurricane~\cite{zahra2020automatic}. 

\textit{\textbf{Arabic Tweet Corpus}} consists of tweets collected during four flood events that took place in different areas of the Arab world (i.e., Jordan, Kuwait, northern Saudi Arabia, and western Saudi Arabia) in 2018~\cite{alharbi2019crisis}. The dataset contains 4,037 labeled tweets with their relevance and information type.

\textit{\textbf{TREC Incident Streams}} dataset has been developed as part of the TREC-IS 2018 evaluation challenge and consists of 19,784 tweets labeled for actionable information identification and assessing the criticality of the information~\cite{mccreadie2019trec}. This dataset is developed based on CrisisLex, CrisisNLP and the data collected using Gnip services. 

\textit{\textbf{Disaster Tweet Corpus 2020}} is a compilation of existing datasets for \textit{disaster event types} classification \cite{Wiegmann2020}.  

\subsection{Modeling Approaches}
\label{ssec:related-work-tools}
For disaster response, typical social media content classification task include {\em(i)} informative vs non-informative, (also referred as related \textit{vs.} not-related, or on-topic \textit{vs.} off-topic), {\em(ii)} fine-grained humanitarian information categories, {\em(iii)} disaster event types, {\em(iv)} damage severity assessment. To address such tasks
classical algorithms have been widely used in developing classifiers 
in the past~\cite{imran2015processing}. However, deep learning algorithms have recently started receiving more attention due to their successful applications in various natural language processing (NLP) and computer vision tasks. For instance, \cite{nguyen2017robust} and \cite{neppallideep} perform comparative experiments between different classical and deep learning algorithms including Support Vector Machines (SVM), Logistic Regression (LR), Random Forests (RF), Recurrent Neural Networks (RNN), and Convolutional Neural Networks (CNN). Their experimental results suggest that CNN outperforms other algorithms. Though in another study, \cite{burel2018crisis} reports that SVM and CNN can provide very competitive results in some cases.
CNNs have also been explored in event type-specific filtering model~\cite{kersten2019}.
Recent successful embedding representations such as Embeddings from Language Models (ELMo)~\cite{peters2018deep}, Bidirectional Encoder Representations from Transformers (BERT)~\cite{devlin2018bert}, and XLNet~\cite{yang2019xlnet} have also been explored for disaster related tweet classification tasks~\cite{jain2019estimating,Wiegmann2020,alam2020standardizing}.
From a modeling perspective, our work is different than previous work in a way that we have used both classical and deep learning algorithms with different transformer-based models, which can serve as a strong baseline for future study.

\begin{table*}[!htb]
	\centering
	\scalebox{0.55}{
		\begin{tabular}{llrcrrrrrrr}
			\toprule
			\multicolumn{1}{l}{\textbf{Event name}} &\multicolumn{1}{l}{\textbf{Event Type}} & \multicolumn{1}{r}{\textbf{Total tweets}} & \multicolumn{1}{c}{\textbf{Date Range}} & \multicolumn{1}{r}{\textbf{Date}} & \multicolumn{1}{r}{\textbf{Location}} & \multicolumn{1}{r}{\textbf{Language}} & \multicolumn{1}{r}{\textbf{Classifier}} & \multicolumn{1}{r}{\textbf{Word Count}} & \multicolumn{1}{r}{\textbf{De-duplication}} & \multicolumn{1}{r}{\textbf{Sampled}} \\ \midrule
			\textbf{2016 Ecuador Earthquake} & Earthquake & 1,756,267 & 2016/04/17 -- 2016/04/18 & 884,783 & 172,067 & 19,988 & 11,269 & 11,251 & 2,007 & 2,000 \\
			\textbf{2016 Canada Wildfires} & Fire & 312,263 & 2016/05/06 -- 2016/05/27 & 312,263 & 66,169 & 66,169 & 5,812 & 5,796 & 2,906 & 2,906 \\
			\textbf{2016 Italy Earthquake} & Earthquake & 224,853 & 2016/08/24 -- 2016/08/29 & 224,853 & 15,440 & 15,440 & 6,624 & 6,606 & 1,458 & 1,458 \\
			\textbf{2016 Kaikoura Earthquake} & Earthquake & 318,256 & 2016/09/01 -- 2016/11/22 & 318,256 & 44,791 & 44,791 & 11,854 & 11,823 & 3,180 & 3,180 \\
			\textbf{2016 Hurricane Matthew} & Hurricane & 1,939,251 & 2016/10/04 -- 2016/10/10 & 82,643 & 36,140 & 36,140 & 10,116 & 10,099 & 2,111 & 2,100 \\
			\textbf{2017 Sri Lanka Floods} & Flood & 40,967 & 2017/05/31 -- 2017/07/03 & 40,967 & 4,267 & 4,267 & 2,594 & 2,594 & 760 & 760 \\
			\textbf{2017 Hurricane Harvey} & Hurricane & 6,384,625 & 2017/08/25 -- 2017/09/01 & 2,919,679 & 1,481,939 & 1,462,934 & 638,611 & 632,814 & 97,034 & 13,229 \\
			\textbf{2017 Hurricane Irma} & Hurricane & 1,641,844 & 2017/09/06 -- 2017/09/17 & 1,266,245 & 563,899 & 552,575 & 113,757 & 113,115 & 29,100 & 13,400 \\
			\textbf{2017 Hurricane Maria} & Hurricane & 2,521,810 & 2017/09/16 -- 2017/10/02 & 1,089,333 & 541,051 & 511,745 & 202,225 & 200,987 & 17,085 & 10,600 \\
			\textbf{2017 Mexico Earthquake} & Earthquake & 361,040 & 2017/09/20 -- 2017/09/23 & 181,977 & 17,717 & 17,331 & 11,662 & 11,649 & 2,563 & 2,563 \\
			\textbf{2018 Maryland Floods} & Flood & 42,088 & 2018/05/28 -- 2018/06/07 & 42,088 & 20,483 & 20,483 & 7,787 & 7,759 & 1,155 & 1,140 \\
			\textbf{2018 Greece Wildfires} & Fire & 180,179 & 2018/07/24 -- 2018/08/18 & 180,179 & 9,278 & 9,278 & 4,896 & 4,888 & 1,815 & 1,815 \\
			\textbf{2018 Kerala Floods} & Flood & 850,962 & 2018/08/17 -- 2018/08/31 & 757,035 & 401,950 & 401,950 & 225,023 & 224,876 & 29,451 & 11,920 \\
			\textbf{2018 Hurricane Florence} & Hurricane & 659,840 & 2018/09/11 -- 2018/09/18 & 483,254 & 318,841 & 318,841 & 38,935 & 38,854 & 13,001 & 9,680 \\
			\textbf{2018 California Wildfires} & Fire & 4,858,128 & 2018/11/10 -- 2018/12/07 & 4,858,128 & 2,239,419 & 2,239,419 & 942,685 & 936,199 & 103,711 & 10,225 \\
			\textbf{2019 Cyclone Idai} & Hurricane & 620,365 & 2019/03/15 -- 2019/04/16 & 620,365 & 47,175 & 44,107 & 26,489 & 26,469 & 5,236 & 5,236 \\
			\textbf{2019 Midwestern U.S. Floods} & Flood & 174,167 & 2019/03/25 -- 2019/04/03 & 174,167 & 96,416 & 96,416 & 19,072 & 19,037 & 3,427 & 3,420 \\
			\textbf{2019 Hurricane Dorian} & Hurricane & 1,849,311 & 2019/08/30 -- 2019/09/02 & 1,849,311 & 993,982 & 993,982 & 137,700 & 136,954 & 18,580 & 11,480 \\
			\textbf{2019 Pakistan Earthquake} & Earthquake & 122,939 & 2019/09/24 -- 2019/09/26 & 122,939 & 34,200 & 34,200 & 16,180 & 16,104 & 2,502 & 2,500 \\\midrule
			\textbf{Total} && \textbf{24,859,155} & \multicolumn{1}{r}{\textbf{}} & \textbf{16,408,465} & \textbf{7,105,224} & \textbf{6,890,056} & \textbf{2,433,291} & \textbf{2,417,874} & \textbf{337,082} & \textbf{109,612} \\\bottomrule
		\end{tabular}
	}
	\caption{Event-wise data distribution, filtering and sampling.}
	\label{table:data_dist_filtering_sampling_details}
\end{table*}

\section{Data Collection, Filtering and Sampling}
\label{sec:dataset}
We used the AIDR system~\cite{imran2014aidr} to collect data from Twitter during 19 disaster events occurred between 2016 and 2019. AIDR is a publicly available system, which uses the Twitter streaming API for data collection.\footnote{\url{http://aidr.qcri.org/}} The data collection was performed using event-specific keywords and hashtags.
In Table~\ref{table:data_dist_filtering_sampling_details}, we list details of the data collection period for each event. 
In total, 24,859,155 tweets were collected from all the events. Data annotation is a costly procedure, therefore, we investigated how to filter and sample data that can maximize the quality of the labeled data. 

\subsection{Data Filtering}
\label{ssec:filtering_and_dampling}
To prepare data for manual annotation, we perform the following filtering steps:

\begin{enumerate}[leftmargin=*]
	\setlength\itemsep{-0.2em}
	\item \textbf{Date-based filtering:} For some events, that data collection period extends beyond the actual event dates. For instance, for Hurricane Florence, our data collection period is from Sep 11, 2018 to Nov 17, 2018 although the hurricane actually dissipated on Sep 18. For this purpose, we restrict the data sampling period to actual event days as reported on their Wikipedia page.
	
	\item \textbf{Location-based filtering:} Since our data collection was based on event-specific keywords (i.e., we did not restrict the data collection to any geographical area), it is likely that a large portion of the collected data come from outside the disaster-hit areas. However, the most useful information for humanitarian organizations is the information that originates from eyewitnesses or people from the disaster-hit areas. Therefore, we discarded all tweets outside the disaster-hit areas by using a geographic filter. The geographic filter uses one of the three fields (i.e., \textit{geo}, \textit{place}, or \textit{user location}) from a tweet. %
	We prioritize the \textit{geo} field, as it comes from the user device as GPS coordinates. If the \textit{geo} field is not available, which is the case for 98\% of the tweets, we use the \textit{place} field. The \textit{place} field comes with a geographical bounding box, which we use to determine whether a tweet is inside or outside an area. As our last option, we use the \textit{user location} field, which is a free-form text provided by the user.
	Next, we use the Nominatim\footnote{\url{http://nominatim.org}} service (an OpenStreetMap\footnote{\url{https://www.openstreetmap.org}} database) to resolve the provided location text into city, state, and country information. The resolved geo-information is then used to filter out tweets which do not belong to a list of locations that we manually curate for each event. A list of target locations for each event is provided in the dataset bundle. 
	
	\item \textbf{Language-based filtering:} We choose to only annotate English tweets due to budget limitations. Therefore, we discard all non-English tweets using the Twitter provided language metadata for a given tweet. It would be interesting to annotate tweets in other languages in future studies. 
	
	\item \textbf{Classifier-based filtering:} After applying the filters mentioned above, the remaining data is still in the order of millions (i.e., $\sim$7 million according to Table~\ref{table:data_dist_filtering_sampling_details}), a large proportion of which might still be irrelevant. To that end, we trained a  Random Forest classifier\footnote{The choice of this algorithm was based on previous studies and its use in practical application \cite{imran2014aidr}. Also because it is computationally simple, which enabled us to classify a large number of tweets in a short amount of time.} using a set of humanitarian categories labeled data reported in \cite{Alam2019}, which consists the classes similar to our annotation task described in the next section. We follow widely used train/dev/test (70/10/20) splits to train and evaluate the model. We prepossessed the tweets before training the classifier, which include removing stop words, URLs, user mentions, and non-ASCII characters. The trained classifier achieved an F1=$76.9\%$, which we used to classify and eliminate all the irrelevant tweets, i.e., tweets classified as not-humanitarian.

	\item \textbf{Word-count-based filtering:} We retain tweets that contain at least three words or hashtags.
	The rationale behind such a choice is that tweets with more tokens tend to provide more information and likely to have additional contextual information useful for responders. %
	URLs and numbers are usually discarded while training a classifier, thus we ignore them while counting the number of tokens for a given tweet. 
	
	\item \textbf{Near-duplicate filtering:} \textcolor{black}{Finally, we apply de-duplication to remove exact and near-duplicate tweets using their textual content. This consists of three steps: {\em (i)} tokenize the tweets to remove URL, user-mentions, and other non-ASCII characters; {\em (ii)} convert the tweets into vectors of uni- and bi-grams with their frequency-based representations, {\em (iii)} compute the cosine similarity between tweets and flag the one as duplicate that exceed a threshold. Since threshold identification is a complex procedure, therefore, we follow the findings in \cite{alam2020standardizing}, where a threshold of $0.75$ is used to flag duplicates.}
\end{enumerate}

\subsection{Sampling}
\label{ssec:sampling}
Although the filtering steps help reduce the total number of tweets significantly while maximizing the information theoretic value of the retained subset, there are still more tweets than our annotation budget. Therefore, in the sampling step, we select $n$ random tweets from each class while also maintaining a fair distribution across classes. In Table \ref{table:data_dist_filtering_sampling_details}, we summarize the details of the data filtering and sampling including total number of tweets initially collected as well as the total number of tweets retained after each filtering and sampling step for each event. In particular, the last column of the table indicates the total number of tweets sampled for annotation for each disaster event.

\begin{table*}[!htb]
	\centering
	\scalebox{0.57}{
		\begin{tabular}{@{}lrrrrrrrrrrrr@{}}
			\toprule
			\multicolumn{1}{@{}l}{\textbf{Event name}} & \multicolumn{1}{r}{\parbox{1.5cm}{\textbf{Caution and advice}}} & \multicolumn{1}{r}{\parbox{1.5cm}{\textbf{Displaced people and evacuations}}} & \multicolumn{1}{r}{\parbox{2cm}{\textbf{Infrastructure and utility damage}}} & \multicolumn{1}{r}{\parbox{1.5cm}{\textbf{Injured or dead people}}} & \multicolumn{1}{r}{\parbox{1.5cm}{\textbf{Missing or found people}}} & \multicolumn{1}{r}{\parbox{1.5cm}{\textbf{Not humanitarian}}} & \multicolumn{1}{r}{\parbox{1.5cm}{\textbf{Don't know or can't judge}}} & \multicolumn{1}{r}{\parbox{1.5cm}{\textbf{Other relevant information}}} & \multicolumn{1}{r}{\parbox{1.5cm}{\textbf{Requests or urgent needs}}} & \multicolumn{1}{r}{\parbox{1.5cm}{\textbf{Rescue volunteering or donation effort}}} & \multicolumn{1}{r}{\parbox{1.5cm}{\textbf{Sympathy and support}}} & \multicolumn{1}{r@{}}{\textbf{Total}} \\ \midrule
			\textbf{2016 Ecuador Earthquake} & 30 & 3 & 70 & 555 & 10 & 23 & 18 & 81 & 91 & 394 & 319 & 1,594 \\
			\textbf{2016 Canada Wildfires} & 106 & 380 & 251 & 4 & - & 79 & 13 & 311 & 20 & 934 & 161 & 2,259 \\
			\textbf{2016 Italy Earthquake} & 10 & 3 & 54 & 174 & 7 & 9 & 10 & 52 & 30 & 312 & 579 & 1,240 \\
			\textbf{2016 Kaikoura Earthquake} & 493 & 87 & 312 & 105 & 3 & 224 & 19 & 311 & 24 & 207 & 432 & 2,217 \\
			\textbf{2016 Hurricane Matthew} & 36 & 38 & 178 & 224 & - & 76 & 5 & 328 & 53 & 326 & 395 & 1,659 \\
			\textbf{2017 Sri Lanka Floods} & 28 & 9 & 17 & 46 & 4 & 20 & 2 & 56 & 34 & 319 & 40 & 575 \\
			\textbf{2017 Hurricane Harvey} & 541 & 688 & 1,217 & 698 & 10 & 410 & 42 & 1,767 & 333 & 2,823 & 635 & 9,164 \\
			\textbf{2017 Hurricane Irma} & 613 & 755 & 1,881 & 894 & 8 & 615 & 60 & 2,358 & 126 & 1,590 & 567 & 9,467 \\
			\textbf{2017 Hurricane Maria} & 220 & 131 & 1,427 & 302 & 11 & 270 & 39 & 1,568 & 711 & 1,977 & 672 & 7,328 \\
			\textbf{2017 Mexico Earthquake} & 35 & 4 & 167 & 254 & 14 & 38 & 3 & 109 & 61 & 984 & 367 & 2,036 \\
			\textbf{2018 Maryland Floods} & 70 & 3 & 79 & 56 & 140 & 77 & 1 & 137 & 1 & 73 & 110 & 747 \\
			\textbf{2018 Greece Wildfires} & 26 & 7 & 38 & 495 & 20 & 74 & 4 & 159 & 25 & 356 & 322 & 1,526 \\
			\textbf{2018 Kerala Floods} & 139 & 56 & 296 & 363 & 7 & 456 & 65 & 955 & 590 & 4,294 & 835 & 8,056 \\
			\textbf{2018 Hurricane Florence} & 1,310 & 637 & 320 & 297 & - & 1,060 & 95 & 636 & 54 & 1,478 & 472 & 6,359 \\
			\textbf{2018 California Wildfires} & 139 & 368 & 422 & 1,946 & 179 & 1,318 & 68 & 1,038 & 79 & 1,415 & 472 & 7,444 \\
			\textbf{2019 Cyclone Idai} & 89 & 57 & 354 & 433 & 19 & 80 & 11 & 407 & 143 & 1,869 & 482 & 3,944 \\
			\textbf{2019 Midwestern U.S. Floods} & 79 & 8 & 140 & 14 & 1 & 389 & 27 & 273 & 46 & 788 & 165 & 1,930 \\
			\textbf{2019 Hurricane Dorian} & 1,369 & 802 & 815 & 60 & 1 & 874 & 46 & 1,444 & 179 & 987 & 1,083 & 7,660 \\
			\textbf{2019 Pakistan Earthquake} & 71 & - & 125 & 401 & 1 & 213 & 32 & 154 & 19 & 152 & 823 & 1,991 \\ \midrule
			\textbf{Total} & \textbf{5,404} & \textbf{4,036} & \textbf{8,163} & \textbf{7,321} & \textbf{435} & \textbf{6,305} & \textbf{560} & \textbf{12,144} & \textbf{2,619} & \textbf{21,278} & \textbf{8,931} & \textbf{77,196} \\ \bottomrule
		\end{tabular}%
	}
	\caption{Distribution of annotations across events and class labels.}
	\label{table:data_dist_annotation}
\end{table*}

\section{Manual Annotations}
\label{sec:manual_annotations}

Since the main purpose of this work is to create a large-scale dataset that can be used to train models that understand the type of humanitarian aid-related information posted in a tweet during disasters, we first define what ``humanitarian aid'' means. For the annotation we opted and redefined the annotation guidelines discussed in \cite{alam2018crisismmd}. 

\textbf{Humanitarian aid:\footnote{\url{https://en.wikipedia.org/wiki/Humanitarian_aid}}} In response to humanitarian crises including natural and human-induced disasters, humanitarian aid involves assisting people who need help. The primary purpose of humanitarian aid is to save lives, reduce suffering, and rebuild affected communities. Among the people in need belong homeless, refugees, and victims of natural disasters, wars, and conflicts who need necessities like food, water, shelter, medical assistance, and damage-free critical infrastructure and utilities such as roads, bridges, power-lines, and communication poles. 

Based on the \textit{Humanitarian aid} definition above, we define each humanitarian information category as follows:\footnote{Note that we also supplemented these definitions by showing example tweets in the instructions.} The annotation task was to assign one of the below labels to a tweet. Though multiple labels can be assigned to a tweet, however, we limited it to one category to reduce the annotation efforts. 
\begin{itemize}[leftmargin=*]
	\setlength\itemsep{-0.10em}
	\item \textbf{Caution and advice:} Reports of warnings issued or lifted, guidance and tips related to the disaster;
	\item \textbf{Sympathy and support:} Tweets with prayers, thoughts, and emotional support;
	\item \textbf{Requests or urgent needs:} Reports of urgent needs or supplies such as food, water, clothing, money, medical supplies or blood;
	\item \textbf{Displaced people and evacuations:} People who have relocated due to the crisis, even for a short time (includes evacuations);
	\item \textbf{Injured or dead people:} Reports of injured or dead people due to the disaster;
	\item \textbf{Missing or found people:} Reports of missing or found people due to the disaster event;
	\item \textbf{Infrastructure and utility damage:} Reports of any type of damage to infrastructure such as buildings, houses, roads, bridges, power lines, communication poles, or vehicles;
	\item \textbf{Rescue, volunteering, or donation effort:} Reports of any type of rescue, volunteering, or donation efforts such as people being transported to safe places, people being evacuated, people receiving medical aid or food, people in shelter facilities, donation of money, or services, etc.;
	\item \textbf{Other relevant information:} If the tweet does not belong to any of the above categories, but it still contains important information useful for humanitarian aid, belong to this category;
	\item \textbf{Not humanitarian:} If the tweet does not convey humanitarian aid-related information;
	\item \textbf{Don't know or can't judge:} If the tweet is irrelevant or cannot be judged due to non-English content.
\end{itemize}

For the manual annotation, we opted to use Amazon Mechanical Turk (AMT) platform. In crowdsourcing, one of the challenges is to find a large number of qualified workers while filtering out low-quality workers or spammers~\cite{chowdhury2014cross}. 
To tackle this problem, a typical approach is to use qualification tests followed by a gold standard evaluation~\cite{chowdhury2015selection}.
We created a qualification test consisting of 10 tweets. To participate in the task, each annotator first needs to pass the qualification test. In order to pass the test, the annotator needs to correctly answer at least 6 out of 10 tweets. The gold standard evaluation is performed at the HIT (i.e., Human Intelligence Task) level.
A HIT consists of 30 tweets and in each HIT there are 10 gold standard tweets (i.e., tweets with known labels) and 20 tweets with unknown labels. These 10 gold standard tweets are selected from a pool of tweets labeled by domain experts. Note that developing a gold standard dataset is another costly procedure in terms of time and money. Therefore, we first randomly selected the tweets from different events by focusing on disaster types such as hurricane, flood, fire and earthquake, and then domain experts manually labeled them. 

The annotator who participates in the HIT needs to read each tweet and assign one of the above labels to complete the task. The participation and completion of a HIT by an annotator are referred to as an assignment. We set the assignment approval criterion to $70\%$, which means an assignment of the HIT will be automatically approved if the annotator correctly labels at least 7 out of 10 gold standard tweets.

For each HIT and the associated tweets, we wanted to have three judgments. As our HIT design consists of 20 tweets with unknown labels and we wanted to automatically approve the HIT, we set the HIT approval criterion to $66\%$. That is, a HIT is approved if the annotators agree on a label for at least 14 out of 20 tweets. In order to approve the label for a tweet, we also set a threshold of $66\%$, which means out of three annotators two of them have to agree on the same label. Since the social media content is highly noisy and categories can be subjective, we choose to use a minimum threshold of $66\%$ for the agreement of the label for each tweet. 

\subsection{Crowdsourcing Results}
\label{ssec:crowdsourcing_results}
In Table~\ref{table:data_dist_filtering_sampling_details}, the last column represents the number of tweets sampled for the annotation (i.e., 109,612 in total). Since in AMT our minimum threshold to accept a HIT was $66\%$, we can expect to acquire agreed labels for a minimum of $66\%$ of the tweets. In Table~\ref{table:data_dist_annotation}, we present the annotated dataset, which consists of class label distribution for each event along with the total number of annotated tweets. In summary, we have ${\sim}70\%$ tweets with agreed labels, which results in 77,196 tweets. 

To compute the annotation agreement we considered the following evaluation measures. 
\begin{enumerate}[leftmargin=*]
	\itemsep-0.1em 
	\item Fleiss kappa: It is a reliability measure that is applicable for any fixed number of annotators annotating categorical labels to a fixed number of items, which can handle two or more categories and annotators~\cite{fleiss2013statistical}. However, it can not handle missing labels, except for excluding them from the computation. 
	\item Average observed agreement: It is an average observed agreement over all pairs of annotators \cite{fleiss2013statistical}.
	\item Majority agreement: We compute the majority at the tweet level and take the average. The reason behind this is that for many tweets the number of annotators vary between three and five, and hence, it is plausible to evaluate the agreement at the tweet level.
	\item Krippendorff’s alpha: It is a measure of agreement that allows two or more annotators and categories~\cite{krippendorff1970estimating}. Additionally, it handles missing labels. 
\end{enumerate}

For the first two methods, we selected three annotations whereas for the last two methods we considered all annotations (i.e., three to five). In Table~\ref{table:data_dist_annotation_agreement}, we present the annotation agreement for all events with different approaches mentioned above. The average agreement score varies $55\%$ to $83\%$. Note that in Kappa measurement value of $0.41$-$0.60$, $0.61$-$0.80$, and $0.81$-$1$ refers to the moderate, substantial, and perfect agreement, respectively~\cite{landis1977measurement}. Based on these measurements we can conclude that our annotation agreement score leads to moderate to the substantial agreement, given the difficult nature of the annotation task.

\begin{table}[h]
	\centering
	\scalebox{0.55}{
		\begin{tabular}{@{}lrrrr@{}}
			\toprule
			\multicolumn{1}{l}{\textbf{Event name}} & \multicolumn{1}{r}{\textbf{Fleiss ($\kappa$)}} & \multicolumn{1}{r}{\textbf{K-$\alpha$}} & \multicolumn{1}{r}{\textbf{Avg obs.}} & \multicolumn{1}{r@{}}{\textbf{Maj agr.}} \\ \midrule
			2016 Ecuador Earthquake & 0.65 & 0.64 & 0.73 & 0.86 \\
			2016 Canada Wildfires & 0.54 & 0.61 & 0.63 & 0.85 \\
			2016 Italy Earthquake & 0.64 & 0.69 & 0.74 & 0.89 \\
			2016 Kaikoura Earthquake & 0.57 & 0.57 & 0.63 & 0.81 \\
			2016 Hurricane Matthew & 0.60 & 0.57 & 0.66 & 0.82 \\
			2017 Sri Lanka Floods & 0.47 & 0.51 & 0.61 & 0.83 \\
			2017 Hurricane Harvey & 0.55 & 0.57 & 0.62 & 0.82 \\
			2017 Hurricane Irma & 0.55 & 0.53 & 0.63 & 0.80 \\
			2017 Hurricane Maria & 0.54 & 0.53 & 0.62 & 0.80 \\
			2017 Mexico Earthquake & 0.57 & 0.62 & 0.67 & 0.86 \\
			2018 Maryland Floods & 0.52 & 0.56 & 0.58 & 0.81 \\
			2018 Greece Wildfires & 0.63 & 0.65 & 0.70 & 0.86 \\
			2018 Kerala Floods & 0.50 & 0.50 & 0.63 & 0.82 \\
			2018 Hurricane Florence & 0.50 & 0.54 & 0.57 & 0.80 \\
			2018 California Wildfires & 0.55 & 0.59 & 0.61 & 0.83 \\
			2019 Cyclone Idai & 0.51 & 0.52 & 0.61 & 0.82 \\
			2019 Midwestern U.S. Floods & 0.48 & 0.50 & 0.58 & 0.80 \\
			2019 Hurricane Dorian & 0.59 & 0.55 & 0.65 & 0.81 \\
			2019 Pakistan Earthquake & 0.55 & 0.61 & 0.65 & 0.85 \\ \midrule
			\textbf{Average} & 0.55 & 0.57 & 0.64 & 0.83 \\ \bottomrule
		\end{tabular}
	}
	\caption{Annotation agreement scores for different events. Metrics: Fleiss $\kappa$, Krippendorff alpha (K-$\alpha$), Average observed agreement (Avg obs.), Average majority agreement (Maj agr.).}
	\label{table:data_dist_annotation_agreement}
\end{table}

In table \ref{tab:annotated_tweet_examples}, we provide a few examples of annotated tweets that show tweet texts clearly demonstrate the labels. 
\begin{table}[h]
	\centering
	\scalebox{0.60}{
		\begin{tabular}{@{}ll@{}}
			\toprule
			\multicolumn{1}{c}{\textbf{Tweet}} & \multicolumn{1}{c}{\textbf{Label}} \\ \midrule
			\begin{tabular}[c]{@{}l@{}}VLEs extending helping hands to provide relief material to affected \\ people in   Keralas Chenganoor, Alapuzha and Idukki districts. \\ They distributed food,   water, clothing, medicine etc to flood people \\ as humanitarian work is still   on. \#KeralaFloodRelief\\  \#KeralaFloods2018 \#OpMadad\end{tabular} & \begin{tabular}[c]{@{}l@{}}Rescue volunteering \\ or donation effort\end{tabular} \\ \midrule
			\begin{tabular}[c]{@{}l@{}}@narendramodi Sir, Chenganoor area is in very dangerous \\ condition..We need more army   assistance there..Please Please \\ help.@PMOIndia \#KeralaFloodRelief\end{tabular} & \begin{tabular}[c]{@{}l@{}}Requests or \\ urgent needs\end{tabular} \\ 
			\midrule
			\begin{tabular}[c]{@{}l@{}}In this difficult period. My prayers for all flood affected people \\ of Kerala. We know Kerala is most beautiful state of india \\ and people of Kerala part of UAE success. Let us extend \\ our hands in support in their difficulties. \\ \#KeralaFloods \#Kerala\_in\_our\_hearts\_\end{tabular} & \begin{tabular}[c]{@{}l@{}}Sympathy \\ and support\end{tabular} \\ 
			\bottomrule
		\end{tabular}
	}
	\caption{Examples of annotated tweets.}
	\label{tab:annotated_tweet_examples}
\end{table}

\subsection{Lexical Analysis and Statistics}
\label{ssec:lexical_analysis}
\begin{figure}[h!]
	\centering
	\includegraphics[width=0.8\linewidth]{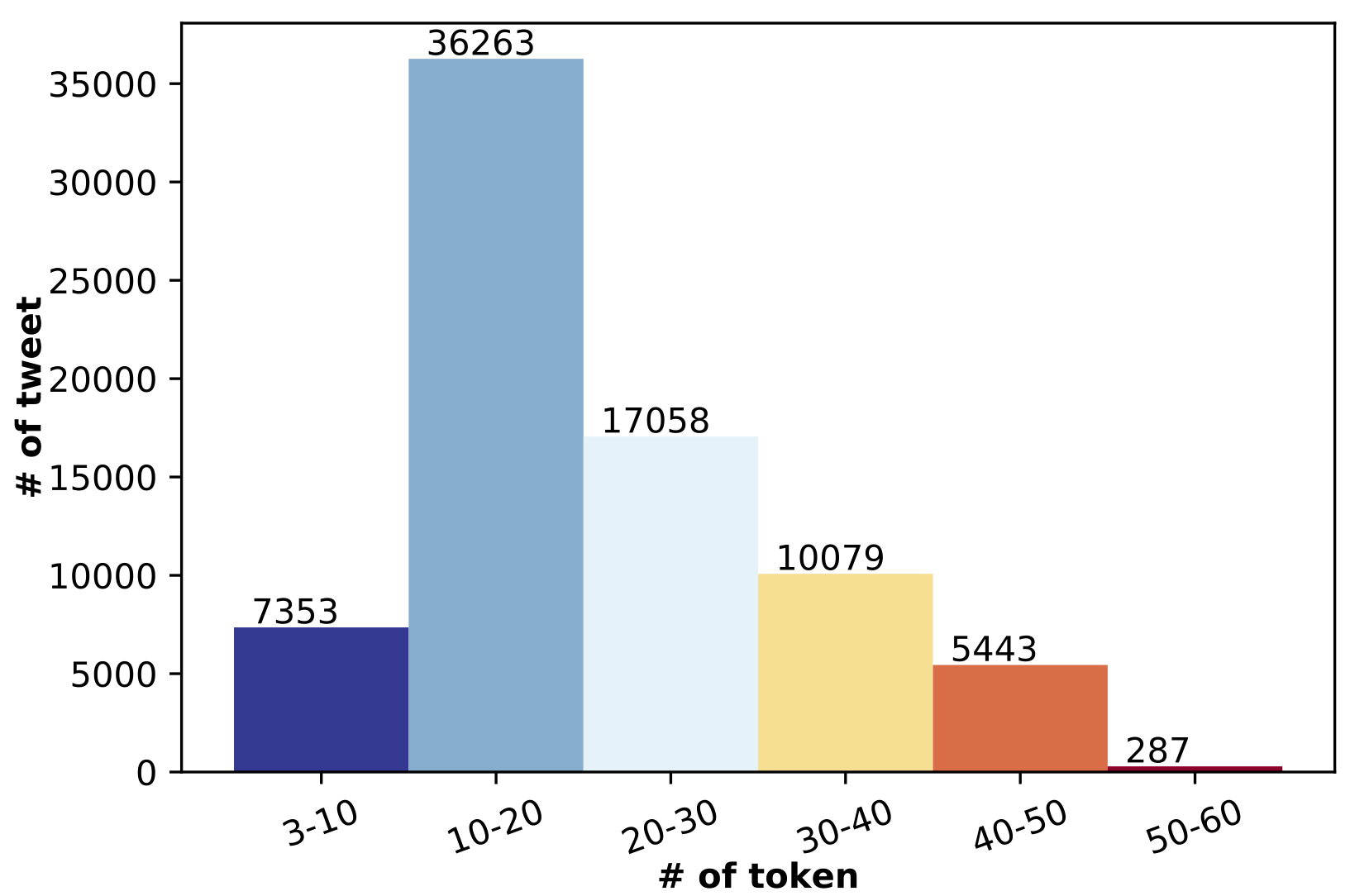}
	\caption{Number of tweet with different lengths in overall dataset.}
	\label{fig:tweet_len_stat}
\end{figure}

To understand the lexical content, we check the number of tokens for each tweet in each event. This information help in understanding the characteristics of the dataset. For example, the maximum number of tokens can help define max sequence length in deep learning-based architectures such as CNN. In Table \ref{tab:tweet_length}, we provide results for each event. The minimum number of tokens is 3 for all the events, therefore, we have not reported that number in the table. From the table, we can observe that for some events, (e.g., Hurricane Maria, Maryland Floods), the max token limits are higher. 
This is because Twitter extended its character limit to 280 from September 2017. 
In Figure \ref{fig:tweet_len_stat}, we provide statistics of the tweet lengths in the overall dataset in different bins. The majority of the tweets is appearing with a range of length 10-20 tokens, second bin in the figure.   

\begin{table}[h]
	\centering
	\scalebox{0.70}{
		\begin{tabular}{@{}lrrr@{}}
			\toprule
			\multicolumn{1}{c}{\textbf{Event name}} & \multicolumn{1}{c}{\textbf{Std.}} & \multicolumn{1}{c}{\textbf{Mean}} & \multicolumn{1}{c}{\textbf{Max}} \\ \midrule
			2016 Ecuador Earthquake & 4.24 & 13.84 & 27 \\
			2016 Canada Wildfires & 4.10 & 14.27 & 28 \\
			2016 Italy Earthquake & 4.44 & 14.11 & 25 \\
			2016 Kaikoura Earthquake & 5.01 & 15.39 & 28 \\
			2016 Hurricane Matthew & 4.51 & 15.76 & 29 \\
			2017 Sri Lanka Floods & 4.15 & 15.93 & 24 \\
			2017 Hurricane Harvey & 4.70 & 15.45 & 31 \\
			2017 Hurricane Irma & 4.89 & 15.47 & 29 \\
			2017 Hurricane Maria & 4.95 & 15.96 & 51 \\
			2017 Mexico Earthquake & 4.64 & 15.49 & 37 \\
			2018 Maryland Floods & 11.06 & 22.75 & 51 \\
			2018 Greece Wildfires & 11.73 & 23.06 & 54 \\
			2018 Kerala Floods & 11.43 & 26.38 & 54 \\
			2018 Hurricane Florence & 10.98 & 25.57 & 55 \\
			2018 California Wildfires & 12.02 & 24.72 & 57 \\
			2019 Cyclone Idai & 11.12 & 28.46 & 53 \\
			2019 Midwestern U.S. Floods & 11.62 & 27.50 & 54 \\
			2019 Hurricane Dorian & 12.14 & 25.73 & 57 \\
			2019 Pakistan Earthquake & 11.71 & 22.80 & 54 \\ \bottomrule
		\end{tabular}
	}
	\caption{Descriptive statistics (i.e., std., max and mean number of token) for each event}
	\label{tab:tweet_length}
\end{table}

\begin{table*}[!htb]
	\centering
	\scalebox{0.56}{
		\begin{tabular}{@{}llrrrrrrrrrrr@{}}
			\toprule
			\multicolumn{1}{@{}l}{\textbf{Event name}} & \multicolumn{1}{c}{\textbf{Data Split}} & \multicolumn{10}{c}{\textbf{Class labels}} & \multicolumn{1}{r@{}}{\textbf{Total}} \\ \midrule
			\textbf{} & \multicolumn{1}{c}{\textbf{}} & \multicolumn{1}{c}{\parbox{1.5cm}{\textbf{Caution and advice}}} & \multicolumn{1}{c}{\parbox{1.5cm}{\textbf{Displaced people and evacuations}}} & \multicolumn{1}{c}{\parbox{2cm}{\textbf{Infrastructure and utility damage}}} & \multicolumn{1}{c}{\parbox{1.5cm}{\textbf{Injured or dead people}}} & \multicolumn{1}{c}{\parbox{1.5cm}{\textbf{Missing or found people}}} & \multicolumn{1}{c}{\parbox{1.5cm}{\textbf{Not humanitarian}}} & \multicolumn{1}{c}{\parbox{1.5cm}{\textbf{Other relevant information}}} & \multicolumn{1}{c}{\parbox{1.5cm}{\textbf{Requests or urgent needs}}} & \multicolumn{1}{c}{\parbox{1.5cm}{\textbf{Rescue volunteering or donation effort}}} & \multicolumn{1}{c}{\parbox{1.5cm}{\textbf{Sympathy and support}}} & \multicolumn{1}{l}{} \\ \midrule
			\multirow{3}{*}{\textbf{2016 Ecuador Earthquake}} & Train & 21 & - & 49 & 388 & - & 16 & 57 & 64 & 276 & 223 & 1,094 \\
			& Dev & 3 & - & 7 & 57 & - & 2 & 8 & 9 & 40 & 33 & 159 \\
			& Test & 6 & - & 14 & 110 & - & 5 & 16 & 18 & 78 & 63 & 310 \\ \midrule
			\multirow{3}{*}{\textbf{2016 Canada Wildfires}} & Train & 74 & 266 & 176 & - & - & 55 & 218 & 14 & 653 & 113 & 1,569 \\
			& Dev & 11 & 39 & 25 & - & - & 8 & 32 & 2 & 95 & 16 & 228 \\
			& Test & 21 & 75 & 50 & - & - & 16 & 61 & 4 & 186 & 32 & 445 \\ \midrule
			\multirow{3}{*}{\textbf{2016 Italy Earthquake}} & Train & - & - & 38 & 122 & - & - & 36 & 21 & 218 & 405 & 840 \\
			& Dev & - & - & 5 & 18 & - & - & 5 & 3 & 32 & 59 & 122 \\
			& Test & - & - & 11 & 34 & - & - & 11 & 6 & 62 & 115 & 239 \\\midrule
			\multirow{3}{*}{\textbf{2016 Kaikoura Earthquake}} & Train & 345 & 61 & 218 & 73 & - & 157 & 218 & 17 & 145 & 302 & 1,536 \\
			& Dev & 50 & 9 & 32 & 11 & - & 23 & 32 & 2 & 21 & 44 & 224 \\
			& Test & 98 & 17 & 62 & 21 & - & 44 & 61 & 5 & 41 & 86 & 435 \\\midrule
			\multirow{3}{*}{\textbf{2016 Hurricane Matthew}} & Train & 25 & 27 & 125 & 157 & - & 53 & 229 & 37 & 228 & 276 & 1,157 \\
			& Dev & 7 & 7 & 35 & 44 & - & 15 & 66 & 11 & 65 & 79 & 329 \\
			& Test & 4 & 4 & 18 & 23 & - & 8 & 33 & 5 & 33 & 40 & 168 \\ \midrule
			\multirow{3}{*}{\textbf{2017 Sri Lanka Floods}} & Train & 20 & - & 12 & 32 & - & 14 & 39 & 24 & 223 & 28 & 392 \\
			& Dev & 3 & - & 2 & 5 & - & 2 & 6 & 3 & 32 & 4 & 57 \\
			& Test & 5 & - & 3 & 9 & - & 4 & 11 & 7 & 64 & 8 & 111 \\ \midrule
			\multirow{3}{*}{\textbf{2017 Hurricane Harvey}} & Train & 379 & 482 & 852 & 488 & - & 287 & 1,237 & 233 & 1,976 & 444 & 6,378 \\
			& Dev & 55 & 70 & 124 & 71 & - & 42 & 180 & 34 & 288 & 65 & 929 \\
			& Test & 107 & 136 & 241 & 139 & - & 81 & 350 & 66 & 559 & 126 & 1,805 \\ \midrule
			\multirow{3}{*}{\textbf{2017 Hurricane Irma}} & Train & 429 & 528 & 1,317 & 626 & - & 430 & 1,651 & 88 & 1,113 & 397 & 6,579 \\
			& Dev & 62 & 77 & 192 & 91 & - & 63 & 240 & 13 & 162 & 58 & 958 \\
			& Test & 122 & 150 & 372 & 177 & - & 122 & 467 & 25 & 315 & 112 & 1,862 \\ \midrule
			\multirow{3}{*}{\textbf{2017 Hurricane Maria}} & Train & 154 & 92 & 999 & 211 & - & 189 & 1,097 & 498 & 1,384 & 470 & 5,094 \\
			& Dev & 22 & 13 & 145 & 31 & - & 28 & 160 & 72 & 202 & 69 & 742 \\
			& Test & 44 & 26 & 283 & 60 & - & 53 & 311 & 141 & 391 & 133 & 1,442 \\ \midrule
			\multirow{3}{*}{\textbf{2017 Mexico Earthquake}} & Train & 24 & - & 117 & 178 & - & 27 & 76 & 43 & 688 & 257 & 1,410 \\
			& Dev & 4 & - & 17 & 26 & - & 4 & 11 & 6 & 100 & 37 & 205 \\
			& Test & 7 & - & 33 & 50 & - & 7 & 22 & 12 & 196 & 73 & 400 \\ \midrule
			\multirow{3}{*}{\textbf{2018 Maryland Floods}} & Train & 49 & - & 55 & 39 & 98 & 54 & 96 & - & 51 & 77 & 519 \\
			& Dev & 7 & - & 8 & 6 & 14 & 8 & 14 & - & 7 & 11 & 75 \\
			& Test & 14 & - & 16 & 11 & 28 & 15 & 27 & - & 15 & 22 & 148 \\ \midrule
			\multirow{3}{*}{\textbf{2018 Greece Wildfires}} & Train & 18 & - & 27 & 346 & 14 & 52 & 111 & 18 & 249 & 225 & 1,060 \\
			& Dev & 3 & - & 4 & 50 & 2 & 8 & 16 & 2 & 36 & 33 & 154 \\
			& Test & 5 & - & 7 & 99 & 4 & 14 & 32 & 5 & 71 & 64 & 301 \\ \midrule
			\multirow{3}{*}{\textbf{2018 Kerala Floods}} & Train & 97 & 39 & 207 & 254 & - & 319 & 669 & 413 & 3,005 & 585 & 5,588 \\
			& Dev & 14 & 6 & 30 & 37 & - & 47 & 97 & 60 & 438 & 85 & 814 \\
			& Test & 28 & 11 & 59 & 72 & - & 90 & 189 & 117 & 851 & 165 & 1,582 \\ \midrule
			\multirow{3}{*}{\textbf{2018 Hurricane Florence}} & Train & 917 & 446 & 224 & 208 & - & 742 & 445 & 38 & 1,034 & 330 & 4,384 \\
			& Dev & 134 & 65 & 33 & 30 & - & 108 & 65 & 5 & 151 & 48 & 639 \\
			& Test & 259 & 126 & 63 & 59 & - & 210 & 126 & 11 & 293 & 94 & 1,241 \\ \midrule
			\multirow{3}{*}{\textbf{2018 California Wildfires}} & Train & 97 & 258 & 295 & 1,362 & 125 & 923 & 727 & 55 & 991 & 330 & 5,163 \\
			& Dev & 14 & 38 & 43 & 199 & 18 & 134 & 106 & 8 & 144 & 48 & 752 \\
			& Test & 28 & 72 & 84 & 385 & 36 & 261 & 205 & 16 & 280 & 94 & 1,461 \\ \midrule
			\multirow{3}{*}{\textbf{2019 Cyclone Idai}} & Train & 62 & 40 & 248 & 303 & 13 & 56 & 285 & 100 & 1,308 & 338 & 2,753 \\
			& Dev & 9 & 6 & 36 & 44 & 2 & 8 & 41 & 15 & 191 & 49 & 401 \\
			& Test & 18 & 11 & 70 & 86 & 4 & 16 & 81 & 28 & 370 & 95 & 779 \\ \midrule
			\multirow{3}{*}{\textbf{2019 Midwestern U.S. Floods}} & Train & 55 & - & 98 & - & - & 272 & 191 & 32 & 552 & 116 & 1,316 \\
			& Dev & 8 & - & 14 & - & - & 40 & 28 & 5 & 80 & 16 & 191 \\
			& Test & 16 & - & 28 & - & - & 77 & 54 & 9 & 156 & 33 & 373 \\ \midrule
			\multirow{3}{*}{\textbf{2019 Hurricane Dorian}} & Train & 958 & 561 & 571 & 42 & - & 612 & 1,011 & 125 & 691 & 758 & 5,329 \\
			& Dev & 140 & 82 & 83 & 6 & - & 89 & 147 & 18 & 101 & 110 & 776 \\
			& Test & 271 & 159 & 161 & 12 & - & 173 & 286 & 36 & 195 & 215 & 1,508 \\ \midrule
			\multirow{3}{*}{\textbf{2019 Pakistan Earthquake}} & Train & 50 & - & 87 & 281 & - & 149 & 108 & 13 & 106 & 576 & 1,370 \\
			& Dev & 7 & - & 13 & 41 & - & 22 & 15 & 2 & 15 & 84 & 199 \\
			& Test & 14 & - & 25 & 79 & - & 42 & 31 & 4 & 31 & 163 & 389 \\ \bottomrule
		\end{tabular} 
	}
	\caption{Event wise data split and distribution of class labels.}
	\label{table:data_split}
\end{table*}

\section{Experiments and Results}
\label{sec:experimental_design}
In this section, we describe the details of our classification experiments and results.
To run the experiments, we split data into training, development, and test sets with a proportion of 70\%, 10\%, and 20\%, respectively. We removed low prevalent classes (i.e., number of tweets with a class label less than 15, e.g., in event \textit{2016 Ecuador Earthquake}, there were only 3 tweets with the class label \textit{Displaced people and evacuations}) in some events from the classification experiments. This approach reduced from 77196 to 76484 tweets for the experiments. 
Event-wise data split and class label distribution are reported in Table \ref{table:data_split}. The event type and combined splits are the merged version from event-wise splits.

We ran classification experiments at three levels: {\em (i)} event level, {\em (ii)} event-type level, and {\em(iii)} all data combined. The purpose of event and event type level experiments is to provide a baseline, which can be used to compare cross event experiments in future studies. For the data splits we first create splits for each event separately. Then, for the event-type experiments, we combine the training, development, and test sets of all the events that belong to the same event type. For example, we combine all training sets of specific earthquake collections into the general earthquake-type training set. Since combining data from multiple events can result in near-duplicate tweets\footnote{This happens when more than one crisis events occur at the same time and same tweets are collected for different events.} across different data splits (i.e., training, development, and test), we applied the same near-duplicate removal approach discussed earlier to eliminate such cases. With this approach, we removed only nine tweets, which leads to having a total of 76,475 tweets in event-type experiments. Similarly, the same duplicate removal approach was applied when combining data from all the events, which also reduced another 9 tweets, resulting 76,466 tweets. 

To measure the performance of each classifier, we use weighted average precision (P), recall (R), and F1-measure (F1). %
The choice of the weighted metric is to factor in the class imbalance problem.

\begin{table*}[!htp]
	\centering
	\scalebox{0.68}{
		\begin{tabular}{@{}lrrrrrrrr@{}}
			\toprule
			\multicolumn{1}{c}{\textbf{Data}} &
			\multicolumn{1}{c}{\textbf{\# Cls}} &
			\multicolumn{1}{c}{\textbf{RF}} &
			\multicolumn{1}{c}{\textbf{SVM}} &
			\multicolumn{1}{c}{\textbf{FastText}} &
			\multicolumn{1}{c}{\textbf{BERT}} &
			\multicolumn{1}{c}{\textbf{D-BERT}} &
			\multicolumn{1}{c}{\textbf{RoBERTa}} &
			\multicolumn{1}{c}{\textbf{XLM-R}} \\ \midrule
			2016 Ecuador Earthquake     & 8  & 0.784 & 0.738 & 0.752 & 0.861 & \textbf{0.872} & \textbf{0.872} & 0.866 \\
			2016 Canada Wildfires       & 8  & 0.726 & 0.738 & 0.726 & \textbf{0.792} & 0.781 & 0.791 & 0.768 \\
			2016 Italy Earthquake       & 6  & 0.799 & 0.822 & 0.821 & 0.871 & 0.878 & \textbf{0.885} & 0.877 \\
			2016 Kaikoura Earthquake    & 9  & 0.660 & 0.693 & 0.658 & \textbf{0.768} & 0.743 & 0.765 & 0.760 \\
			2016 Hurricane Matthew      & 9  & 0.742 & 0.700 & 0.704 & 0.786 & 0.780 & \textbf{0.815} & 0.784 \\
			2017 Sri Lanka Floods       & 8  & 0.613 & 0.611 & 0.575 & 0.703 & 0.763 & 0.727 & \textbf{0.798} \\
			2017 Hurricane Harvey       & 9  & 0.719 & 0.713 & 0.718 & 0.759 & 0.743 & \textbf{0.763} & 0.761 \\
			2017 Hurricane Irma         & 9  & 0.693 & 0.695 & 0.694 & 0.722 & 0.723 & \textbf{0.730} & 0.717 \\
			2017 Hurricane Maria        & 9  & 0.682 & 0.682 & 0.688 & 0.715 & 0.722 & \textbf{0.727} & 0.723 \\
			2017 Mexico Earthquake      & 8  & 0.800 & 0.789 & 0.797 & 0.845 & 0.854 & \textbf{0.863} & 0.847 \\
			2018 Maryland Floods        & 8  & 0.554 & 0.620 & 0.621 & 0.697 & 0.734 & 0.760 & \textbf{0.798} \\
			2018 Greece Wildfires       & 9  & 0.678 & 0.694 & 0.667 & \textbf{0.788} & 0.739 & 0.783 & 0.783 \\
			2018 Kerala Floods          & 9  & 0.670 & 0.694 & 0.714 & 0.732 & 0.732 & 0.745 & \textbf{0.746} \\
			2018 Hurricane Florence     & 9  & 0.731 & 0.717 & 0.735 & 0.768 & 0.773 & \textbf{0.780} & 0.765 \\
			2018 California Wildfires   & 10 & 0.676 & 0.696 & 0.713 & 0.760 & \textbf{0.767} & 0.764 & 0.757 \\
			2019 Cyclone Idai           & 10 & 0.680 & 0.730 & 0.707 & 0.790 & 0.779 & \textbf{0.796} & 0.793 \\
			2019 Midwestern U.S. Floods & 7  & 0.643 & 0.632 & 0.624 & 0.702 & 0.706 & \textbf{0.764} & 0.726 \\
			2019 Hurricane Dorian       & 9  & 0.688 & 0.663 & \textbf{0.693} & 0.691 & 0.691 & 0.686 & 0.691 \\
			2019 Pakistan Earthquake    & 8  & 0.753 & 0.766 & 0.787 & 0.820 & 0.822 & \textbf{0.834} & 0.827 \\\midrule
			Earthquake                  & 9  & 0.766 & 0.783 & 0.789 & 0.833 & \textbf{0.839} & 0.836 & 0.837 \\
			Fire                        & 10 & 0.685 & 0.717 & 0.727 & 0.771 & 0.771 & \textbf{0.787} & 0.779 \\
			Flood                       & 10 & 0.653 & 0.693 & 0.704 & 0.749 & 0.734 & \textbf{0.758} & 0.755 \\
			Hurricane                   & 10 & 0.702 & 0.716 & 0.730 & 0.740 & \textbf{0.742} & 0.741 & 0.739 \\\midrule
			All                         & 10 & 0.707 & 0.731 & 0.744 & 0.758 & 0.758 & \textbf{0.760} & 0.758 \\ \midrule
			\textbf{Average}            &    & \textbf{0.700} & \textbf{0.710} & \textbf{0.712} & \textbf{0.768} & \textbf{0.769} & \textbf{0.781} & \textbf{0.777} \\ \bottomrule
		\end{tabular}
	}
	\caption{Classification results (weighted F1) for events, event-type and combined (All) dataset. Cls: Number of class labels, XLM-R: XLM-RoBERTa, D-BERT: DistilBERT, Best results are highlighted with \textbf{bold} form.}
	\label{tab:classification_results}
\end{table*}
\subsection{Preprocessing}
\label{ssec:preprocessing_data}
Tweet text consists of many symbols, emoticons, and invisible characters. Therefore, we preprocess them before using in model training and classification experiments. The preprocessing part includes removal of stop words, non-ASCII characters, punctuations (replaced with whitespace), numbers, URLs, and hashtag signs.

\subsection{Models}
\label{ssec:classificaiton_algo}
For this study, we focus on multiclass classification experiments using both classical and deep learning algorithms discussed below. 
As for the classical models, we used the two most popular algorithms {\em (i)} Random Forest (RF) ~\cite{breiman2001random}, and {\em (ii)} Support Vector Machines (SVM) ~\cite{platt1998sequential}.

As deep learning algorithms, we used 
FastText~\citep{joulin2017bag}
and transformer-based models such as BERT \cite{devlin2018bert}, RoBERTa \citep{liu2019roberta}, XLM-RoBERTa~\citep{conneau2019unsupervised} and DistilBERT \cite{sanh2019distilbert}. The reason to choose XLM-RoBERTa is that some tweets can have mix-language (e.g., English tweets with some French words) and we wanted to see how model performs given that it is a multilingual model. 

\subsection{Classification Experiments}
\label{ssec:classificaiton_exp}
To train the classifiers using the aforementioned \textit{classical algorithms}, we converted the preprocessed tweets into bag-of-$n$-gram vectors weighted with logarithmic term frequencies (tf) multiplied with inverse document frequencies (idf). Since contextual information, such as $n$-grams, are useful for classification, we extracted unigram, bigram, and tri-gram features. 
For both SVM and RF we use grid search to optimize the parameters.

For FastText, we used pre-trained embeddings trained on Common Crawl\footnote{\url{https://fasttext.cc/docs/en/crawl-vectors.html}} and default hyperparameter settings.

For transformer-based models, we use the Transformer Toolkit~\citep{Wolf2019HuggingFacesTS} and fine-tune each model using the settings as described in \citep{devlin2018bert} with a task-specific layer on top of the transformer model. Due to the instability of the pre-trained models as reported by~\cite{devlin2018bert}, we do 10 runs of each experiment using different random seeds and choose the model that performs the best on the development set. For training the BERT model for each event, event-type and all combined dataset, we use a batch size of 32, learning rate of 2e-5, maximum sequence length 128, and fine tune 10 epochs with the `categorical cross-entropy' as the loss function. 

\subsection{Results}
\label{ssec:baseline_results_bert}
In Table~\ref{tab:classification_results}, we report the classification results (weighted F1) for each event, event type and combined dataset with all models. 
The column \textit{\# Cls} reports number of class labels available after removing low prevalent class labels. Between classical algorithms, overall, the performance of SVM is better than RF.

The comparison between SVM and FastText, the performances are quite close for many events, and event types. The transformer based models are outperforming across events, event types, and combined dataset.

The comparison of different transformer-based models entails that DistilBERT shows similar performance as opposed to BERT model, therefore, the use of DistilBERT in real applications might be a reasonable choice. RoBERTa and XLM-RoBERTa are outperforming BERT and DistilBERT, which comes with the cost of their large number of parameters. In terms of comparing monolingual (RoBERTa) vs. multilingual (XLM-RoBERTa) version of RoBERTa the gain with monolingual training is higher. 
In terms of comparing event type and all data, for the earthquake event type, we attain higher performance as the classifier is trained and evaluated on nine class labels as opposed to ten class labels for other event types and the combined dataset.

\section{Discussions}
\label{sec:discussion_future_works}
There has been significant progress in crisis informatics research in the last several years due to the growing interest in the community and publicly available resources. 
One of the major interests is analyzing social media data and finding actionable information to facilitate humanitarian organizations. Models have been developed using publicly available datasets to streamline this process. Currently publicly available datasets are limited in different respects such as duplicates, and no fixed test set for the evaluation. These issues make it difficult to understand whether one approach outperforms another.

While closely inspecting the available datasets, we observed that there are duplicates and near-duplicates, which results in misleading performance figures. Another major limitation is that the reported results are not comparable because there is no fixed test set for the evaluation. That is, each set of results has been reported on its own data split, which makes it difficult to understand whether one approach outperforms another. 
To address such limitations and advance the crisis informatics research, in this study, we report our efforts to develop the largest-to-date Twitter dataset (i.e., ${\sim}$77,000 tweets) focusing on humanitarian response tasks. 
While developing the dataset we carefully designed a unique \textit{data filtering and sampling pipeline}, which ensured the following characteristics: {\em(i)} Four major disaster types, {\em ii)} disasters have occurred in different parts of the world at different times, {\em(iii)} selected samples are from disasters that created impact on land and caused major damage, {\em(iv)} content language is English, {\em(v)} data sampling is based on an in-house classifier to eliminate irrelevant content, {\em(vi)} there are at least three words in a tweet, {\em (vii)} exact and near-duplicates are removed (even across events for the combined dataset), and {\em (viii)} moderate to substantial agreement inter-annotator agreement.

While using AMT for annotation, we ensured higher quality annotations by putting in place a qualification test, including gold-standard evaluation inside the tasks, and requiring an agreement score of minimum $66\%$. 

To allow for accurate performance comparison and reporting across future studies, we make the data splits publicly available. We have conducted experiments using different classification algorithms on these data splits to provide baseline results for future research. In total, our experimental setup consists of more than 1000 experiments. Event-wise classification results can be useful to realize within- and across-event experimental comparisons. Whereas, the event-type results can be helpful to develop a generalized event-type-based model and compare it with new approaches. Similarly, a more generalized classifier can be developed using the combined dataset and our provided results can be helpful in future experimental comparison. 

\section{Conclusions}
\label{sec:conclutions}
The information available on social media has been widely used by humanitarian organizations at times of a disaster, which has been posted during an ongoing crisis event. However, most of these posts are not useful or relevant and need to be filtered out to have a concise summary. Besides, fine-grained analysis and understanding are also necessary to take actionable decisions. 
Such fine-grained analysis could be a report of ``infrastructure or utility damage,'' ``urgent needs,'' and so on. 
This requires having a classifier that can categorize such information. 
Our study focused on creating a dataset, which can be used to train a classifier and to categorize such information useful for actionable decisions. To this end, HumAID is the largest dataset, which will be publicly available for the research community. We also provide classification benchmark results, which can used to compare in future studies.

\section*{Ethics and Broader Impact}
We collected tweets from Twitter using Twitter streaming API by following its terms of service. The annotated dataset can be used to develop a model for humanitarian response tasks. We release the dataset by maintaining Twitter data redistribution policy.

{
\small
\bibliography{bib/main}
}

\end{document}